\documentclass[11pt,a4paper]{article}
\usepackage[hyperref]{acl2021}
\usepackage{times}
\usepackage{latexsym}

\usepackage{microtype}
\usepackage{graphicx}

\aclfinalcopy % Uncomment this line for the final submission
\title{ExcavatorCovid: Extracting Events and Relations from Text Corpora for Temporal and Causal Analysis for COVID-19}

\author{Bonan Min, Benjamin Rozonoyer, Haoling Qiu, Alexander Zamanian, Jessica MacBride \\
  Raytheon BBN Technologies, Cambridge, Massachusetts\\
  {\tt bonan.min@raytheon.com}}

\date{}

\begin{document}
\maketitle
\begin{abstract}
Timely responses from policy makers to mitigate the impact of the COVID-19 pandemic rely on a comprehensive grasp of events, their causes, and their impacts. These events are reported at such a speed and scale as to be overwhelming. 
In this paper, we present ExcavatorCovid, a machine reading system that ingests open-source text documents (e.g., news and scientific publications), extracts COVID-19 related events and relations between them, and builds a Temporal and Causal Analysis Graph (TCAG). Excavator will help government agencies alleviate the information overload, understand likely downstream effects of political and economic decisions and events related to the pandemic, and respond in a timely manner to mitigate the impact of COVID-19. We expect the utility of Excavator to outlive the COVID-19 pandemic: analysts and decision makers will be empowered by Excavator to better understand and solve complex problems in the future. An interactive TCAG visualization is available at \url{http://afrl402.bbn.com:5050/index.html}. We also released a demonstration video at \url{https://vimeo.com/528619007}.

\end{abstract}

\section{Introduction}

Timely responses from policy makers to mitigate the impact of the COVID-19 pandemic rely on a comprehensive grasp of events, their causes, and their impacts. Since the beginning of the COVID-19 pandemic, an enormous amount of articles are being published every day, that report many events~\footnote{We define an event as any occurrence, action, process or state of affairs, following ~\cite{ogorman-etal-2016-richer}.} related to COVID as well as studies related to COVID. It is very difficult, if not impossible, to keep track of these developing events or to get a comprehensive overview of the temporal and causal dynamics underlying these events. 

To aid the policy makers in overcoming the information overload, we developed ExcavatorCovid (or Excavator for short), a system that will ingest open-source text sources (e.g., news articles and scientific publications), extract COVID-19 related events and relations between them, and build a Temporal and Causal Analysis Graph (TCAG). Excavator combines the following NLP techniques:
% \vspace{-4mm}
\begin{itemize}
\vspace{-2mm}
    \item Extracting events (\S \ref{sec:events}) for types in our comprehensive COVID-19 event taxonomy (\S \ref{sec:taxonomy}). Each event will have time and location if available in text, allowing analyses targeted at specific times or geographic regions of interest.
    \vspace{-6mm}
    \item Extracting three types of temporal and causal relations (\S \ref{sec:relations}) between pairs of events.
    \vspace{-2mm}
    \item Constructing a TCAG (\S \ref{sec:tcag}) by assembling all events and relations, to provide a comprehensive overview of the events related to COVID-19 as well as their causes and impacts.
    \vspace{-2mm}
    \item Supporting trend and correlation analysis of events, via visualizing event popularity time series  (\S ~\ref{event_timeline}) in the TCAG visualization. 
\end{itemize}
    % \vspace{-3mm}

% Excavator is designed to accept streaming data for real-time updates to the TCAG, therefore, it is also possible to provide a timely overview of the development events and how they unfold. We leave that as future work.
Excavator produces a TCAG that is in a machine-readable JSON format and is also human-understandable (visualized via a web-based interactive User Interface), to support varied analytical and decision making needs. We hope that Excavator will aid government agencies in efforts to understand likely downstream effects of political and economic decisions and events related to the pandemic, and respond in a timely manner to mitigate the impact of COVID-19. The benefit of Excavator is realized through a comprehensive visualization of events and how they affect each other. We expect the utility of Excavator to outlive the COVID-19 pandemic: analysts and decision makers will be empowered by Excavator to better understand and solve complex problems in the future. 

\begin{figure*}
  \centering 
  \includegraphics[scale=0.375]{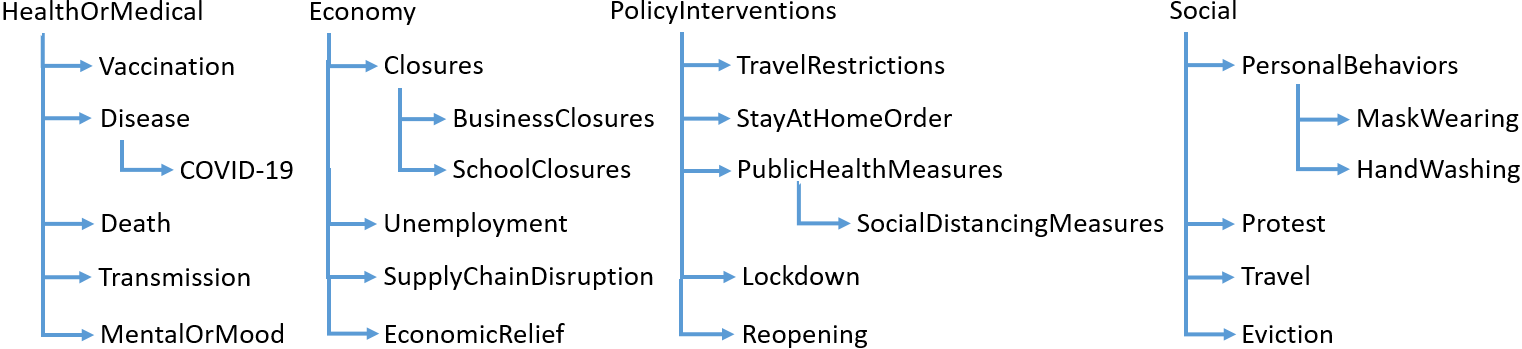}
  \vspace{-1mm}
  \caption{An partial illustration of the COVID-19 event taxonomy.}
    \vspace{-3mm}
  \label{fig:taxonomy}
\end{figure*}
    % \vspace{-3mm}

We first present our COVID-19 event taxonomy, and then we present details about event extraction, causal and temporal relation extraction, measuring event popularity using news text as ``quantitative data'', and the approach for constructing a TCAG. We then describe the system demonstration, present a quantitative analysis of the extractions, and conclude with recommended use cases. 

\section{Building a COVID-19 Event Taxonomy}
\label{sec:taxonomy}

COVID-19 affects many aspects of our political, economic, and personal lives. A comprehensive analysis requires an event taxonomy that categorizes the events related to COVID-19 in many sectors and domains. We developed a COVID-19 event taxonomy using a hybrid approach of manual curation with automated support: first, we run Stanza~\cite{qi2020stanza} on a large sample (10\%) of the Aylien coronavirus news dataset (\S ~\ref{sec:demo}) to tag verb and noun phrases that are likely to trigger events. Second, we represent each phrase as the average of the BERT~\cite{devlin-etal-2019-bert} contextualized embedding vectors of the subwords within each phrase, and then run committee-based clustering~\cite{pantel2002document} over the vector representations of the phrases to discover salient clusters. Finally, we review the frequently appearing clusters and define event types related to COVID-19. 

The event taxonomy includes 76 event types and a short description of each type. Figure~\ref{fig:taxonomy} illustrates several branches of the event taxonomy (the complete taxonomy will be publicly available via \url{github.com}). The events come from a wide range of domains. We also manually added the hyponymy relation via {\it is\_a} links (e.g., COVID-19 {\it is\_a} \{Virus, Disease\}) between pairs of event types.

\section{Extracting Events}
\label{sec:events}

\begin{figure}
  \centering 
  \includegraphics[scale=0.45]{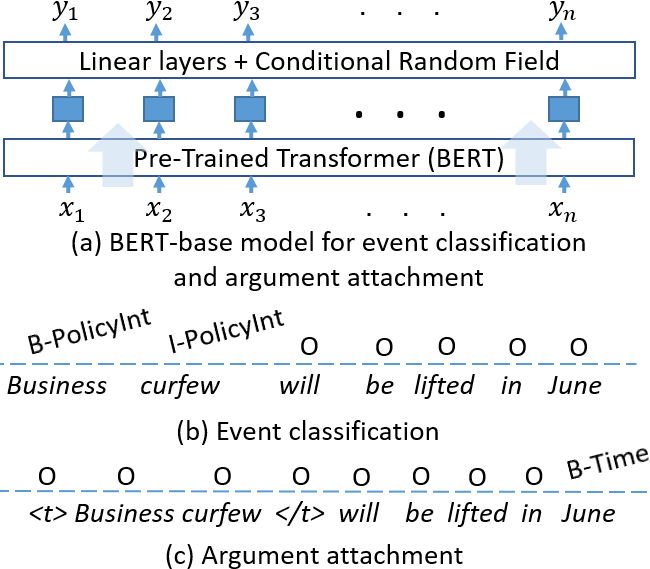}
  \vspace{-7mm}
  \caption{The BERT-based sequence tagging model for event classification and argument extraction. Figure (a) shows the architecture of the model, which takes a sequence of words $x_1, x_2, ..., x_n$ as input and outputs a sequence of tags $y_1, y_2, ..., y_n$. Figure (b) and (c) shows an example for each of the two stages. ``PolicyInt'' is short for ``PolicyIntervention''.}
    \vspace{-5mm}
  \label{fig:nn_events}
\end{figure}

We developed a neural network model for extracting events defined in the COVID-19 event taxonomy (the {\it event classification} stage) and extracting the location and time arguments (the {\it event argument extraction} stage), if they are mentioned in text, for each event mention. The structured representation (events with location and/or time) enables analyses of events targeting a specific time or location. Both stages use a BERT-based sequence tagging model. Figure~\ref{fig:nn_events}(a) shows the model architecture. Given a sequence of tokens as input, the model extracts a sequence of tags, one per each token. We use the commonly used Begin-Inside-Outside (BIO) tags for both event types and event argument role types for the event classification and argument attachment tasks respectively. 

\vspace{-2mm}
\paragraph{Event classification:} a sequence tagging model is trained to predict BIO tags of event types such that it identifies the event trigger span as well as the event type. Figure~\ref{fig:nn_events}(b) shows an example.
\vspace{-2mm}
\paragraph{Event argument extraction:} similarly, another sequence tagging model is trained to predict BIO tags of argument role types, such that it identifies token spans of event arguments as well as their argument role types, with respect to a trigger has already been identified in the {\it event classification} stage and marked in the input sentence in ``$<t>...</t>$''. Figure~\ref{fig:nn_events}(c) shows an example.

We run these two models in a pipeline: the event classification model is applied first to find event triggers and classify their types, then the event argument extraction model is applied to find location and time arguments for each event mention.
\vspace{-2mm}

\paragraph{Training data curation.} 
We apply our prior work on rapid customization for event extraction~\cite{chan-etal-2019-rapid} to curate a dataset for training the event classification model. Our developer spent about 13 minutes per event type to find, expand, and filter potential event triggers in a held-out 10\% of the Aylien coronavirus news corpus. The statistics of the curated data set are shown in Table~\ref{table:freq_events} (we only show the top-10 most frequent event types for brevity). In total, there are 11814 mentions in 7159 sentences. We plan to make this dataset available via \url{github.com}.

To train the argument extraction model, we use the related event-argument annotation from the ACE 2005 dataset~\cite{doddington2004automatic}. We focus on location and time arguments~\footnote{For example, {\it Place} and {\it Time} event argument roles in ACE can be used to train an argument-role model to extract location and time arguments, respectively.} and ignore other roles. At decoding time, after extracting the argument mentions for events, we apply the AWAKE~\cite{boschee2014researching} entity linking system to resolve each location argument to a canonical geolocation, and use SERIF~\cite{boschee2005automatic} to resolve each time argument to a canonical time and then convert it to the month level. This allows us to perform analyses of events targeting a specific geolocation or month of interest.

\begin{table}[t]
\centering
\scalebox{0.75}{
%\begin{small}
\begin{tabular}{|c|c||c|c|}
\hline
{\bf Type} & {\bf Counts} & {\bf Type} & {\bf Counts} \\
\hline
COVID-19 & 2114 & SocialDistancingMeasures & 412 \\
\hline
Virus & 1028 & TravelRestrictions & 403 \\
\hline
Pandemic & 596 & Disease & 378 \\
\hline
Unemployment & 506 & Death & 355 \\
\hline
Shortage & 502 & Lockdown & 321 \\
\hline
\end{tabular}}
\vspace{-2mm}
\caption{Top-10 frequent events in the training dataset. }
\vspace{-2mm}
\label{table:freq_events}
\end{table}
  \vspace{-2mm}

\section{Extracting Temporal and Causal Relations}
\label{sec:relations}

We develop two approaches for extracting temporal and causal relations: a pattern-based approach and a neural network model. 
% Both approaches are tuned to extract relations at $>0.7$ accuracy. 
We take the union of the outputs from both approaches to maximize recall. The list of causal and temporal relations extracted by the systems is shown in Table ~\ref{table:relations}. Our extractors extract relations at the subtype level. However, we decided to merge the subtypes into types because (a) a user survey shows that users prefer to have a simplified definition of causality that only includes ``event X causes (positively impacts) event Y" and ``X mitigates (reduces/prevents) Y", because finer-grained distinctions at sub-type level are difficult and less useful, and (b) merging the subtypes into types improves accuracy to near or above 0.8 as shown in Table ~\ref{tab:extracted_relations}, comparing to 0.7 at the sub-type level due to extraction approaches struggling to differentiate between the sub-types.

\begin{table}[t]
\centering
\scalebox{0.75}{
%\begin{small}
\begin{tabular}{|c|c|c|}
\hline
{\bf Type} & {\bf Subtype} & {\bf Definition} \\
\hline
Causes & Cause & Y happens because of X. \\
& Catalyst & If X, intensity of Y increases. \\
& Precondition & X must have occured for Y to happen. \\
\hline
Mitigates & Mitigation & If X,  intensity of Y decreases. \\
& Preventative & If X happens, Y can’t happen.  \\
\hline
Before & Before/after & X happens before/after Y.   \\
\hline
\end{tabular}}
\vspace{-2mm}
\caption{Causal and temporal relations between event X and Y. }
\vspace{-6mm}
\label{table:relations}
\end{table}

\paragraph{Pattern-based relation extraction.} We applied the temporal and causal relation extraction patterns from LearnIt~\cite{min2020learnit}. A pattern is either a lexical pattern, which is a sequence of words between a pair of events, e.g.,``X leads to Y'' ~\footnote{X and Y refer to the left and right arguments of a relation.}, or a proposition
pattern, which is the (nested) predicate-argument structure that connects the pair of events. For example, ``verb:cause[subject=X] [object=Y]'' is the proposition counterpart of the lexical pattern ``X causes Y''.

\begin{figure}[h]
  \centering 
  \includegraphics[scale=0.375]{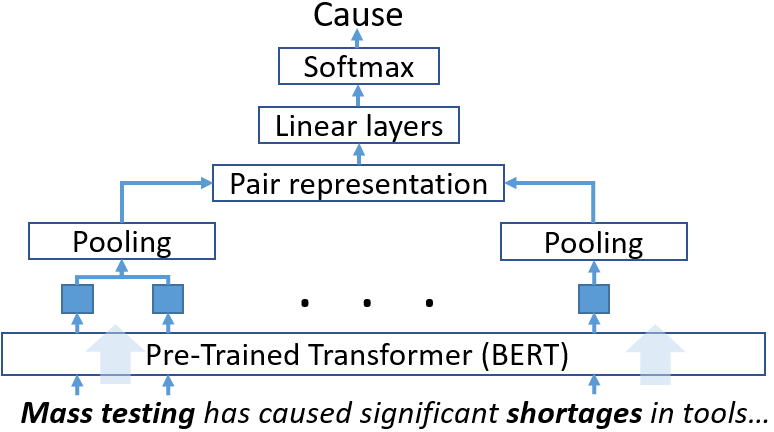}
  \vspace{-6mm}
  \caption{The neural model for causal and temporal relation extraction. }
  \vspace{-3mm}
  \label{fig:nn_relation}
\end{figure}

\paragraph{Neural relation extraction.} We developed a mention pooling ~\cite{baldini-soares-etal-2019-matching} neural model for causal and temporal relation extraction. Figure~\ref{fig:nn_relation} shows the model architecture. Taking a sentence in which a pair of event mention spans are marked as input, the model first encodes the sentence with BERT ~\cite{devlin-etal-2019-bert} ~\footnote{The BERT-Base model is used.}. For each of the left and right event mentions, it then uses average pooling over the BERT contextualized vectors of the words in the span to obtain fixed-dimension vectors $V_1$ and $V_2$ as the span representations. It then concatenates the input embeddings $V_1$ and $V_2$ with the element-wise difference $|V_1-V_2|$ to generate the pair representation $V=(V_1, V_2, |V_1 - V_2|)$. $V$ is passed into a linear layer followed by a softmax layer to make the relation prediction. The model is trained with a blended dataset consisting of the Entities, Events, Simple and Complex Cause Assertion Annotation datasets~\footnote{The catalog IDs of the LDC datasets are LDC2019E48, LDC2019E61, LDC2019E70, LDC2019E82, LDC2019E83.} released by LDC~\footnote{www.ldc.upenn.edu}, and 1.5K temporal relation instances generated by applying the LearnIt temporal relation extraction patterns to 10,000 sampled  Gigaword~\cite{parker2011english} articles.

\vspace{-2mm}

\section{Constructing a TCAG}
\label{sec:tcag}

We aggregate all extracted events and causal and temporal relations across the corpus to construct a TCAG. The TCAG is visualized in the interactive visualization, in which each node is an event type and each edge is a causal or temporal relation~\footnote{{\it is\_a} relations are also added as dashed edges in the TCAG.}.

We use a simple approach to aggregate events: by default, all event mentions sharing the same type are grouped into a single node named by the type; we resort to the UI to allow the user to selectively focus on a specific location and/or time, such that the UI will only show a TCAG involving event mentions and causal relations between pairs of events for the location and/or time of interest. 
% The TCAG is machine-readable in an open JSON format. 

\section{Measuring Event Popularity through Time}
\label{event_timeline}

The TCAG only provides a qualitative analysis of the temporal and causal relations between the COVID-related events. It will be more informative if we can measure the popularity of events through time to enable {\bf trend analysis} (e.g., {\it does lockdown go up or down between January and May, 2020?}) and {\bf correlation analysis} (e.g., {\it will a stricter lockdown improve or deteriorate the economy?}).

In order to support these analyses, we produce a timeseries of a popularity score for each event type over time (a.k.a., event timeline). Extending our prior work ~\cite{min-zhao-2019-measure}, we define the popularity score for event type $e$ at time $t$ as:
        \vspace{-3mm}
        \[
        Popularity(e)_t=\frac{1}{T}\sum_{t'\in [t-\frac{T}{2},t+\frac{T}{2}]}\frac{N_{e,t'}}{M_{t'}} 
                \vspace{-3mm}
        \]
in which $N_{e,t}$ is the frequency of event $e$ at month $t$. We calculate the moving average centered at each $t$ with a sliding window of $T=3$ months to reduce noise. $M_t$ is $1/500$ of the total number of articles published in month $t$. The raw event frequency counts can be inflated due to the increasing level of media activity. Therefore, we divide the raw counts by $M_t$ to normalize the counts so that they are comparable across different months. 

\vspace{-1mm}
\section{System Demonstration}
\label{sec:demo}

\vspace{-1mm}
\paragraph{Datasets.}
We run Excavator on the following two corpora to produce a TCAG for COVID-19: the first corpus is 1.2 million articles~\footnote{These articles do not overlap with the held-out set for training data curation.} from the Aylien Coronavirus News Dataset~\footnote{{\url https://aylien.com/blog/free-coronavirus-news-dataset}}, which contains 1.6 million COVID-related articles published between November 2019 and July 2020 that are from $\sim$440 news sources. We only kept the articles that are published between January and May 2020, since the corpus contains fewer articles in other months. The second corpus is the COVID-19 Open Research Dataset~\cite{wang-etal-2020-cord}. It contains coronavirus-related research from PubMed's PMC corpus, a corpus maintained by the WHO, and bioRxiv and medRxiv pre-prints. As of 11/08/2020, it contains over 300,000 scholarly articles. 

We combine these two corpora because news and research articles are complementary: news are rich in real-world events and are up to date, while analytical articles contain more causal relationships. Therefore, combining them is likely to lead to a more comprehensive analysis and new insights. 

\vspace{-2mm}
\paragraph{Overall statistics of extractions.} 

Excavator extracted 6.2 million event mentions of 59 types. Table~\ref{fig:event_counts} shows the event types that appear more than 50,000 times. We randomly sampled 100 event mentions, manually reviewed them, and found that the extracted events are 83\% accurate. Excavator extracted 226,176 causal and temporal relations from the two corpora. A summary of the extracted relations and their precision~\footnote{Estimated by manually reviewing 40 instances per type} are shown in Table~\ref{tab:extracted_relations}.

\begin{table}[t]
\centering
\scalebox{0.8}{
%\begin{small}
\begin{tabular}{|c|c||c|c|}
\hline
{\bf Type} & {\bf Counts} & {\bf Type} & {\bf Counts} \\
\hline
 COVID-19	& 2772.3 & Travel & 111.8 \\
\hline
 Death	& 730.0 &  FearOrPanic & 94.6 \\
\hline
 Pandemic & 689.2 &  Closures & 92.3 \\
\hline
 Lockdown& 417.2 &  TravelRestrictions & 76.9 \\
\hline
 Isolation* & 195.4 &  Shortage & 68.3 \\
\hline
 DiseaseSpread & 145.4 &  Conflict & 55.5 \\
\hline
 Testing & 130.7 &  Virus & 54.8 \\
\hline
 Treatment & 112.8 &  Symptom & 54.0 \\
\hline

\end{tabular}}
\vspace{-2mm}
  \caption{Frequent events extracted from the corpora (ranked by frequency reversely; numbers are in thousands). *Isolation refers to IsolationOrConfinement.}
  \vspace{-2mm}
  \label{fig:event_counts}
\end{table}

\vspace{-4mm}
\begin{table}[t]
\centering
\scalebox{0.9}{
\begin{tabular}{|c|c|c|c|}
\hline
{\bf Type} & {\bf Count} & {\bf Precision} \\ 
\hline
Causes & 193,694 & 0.78 \\
Mitigates & 30,452 & 0.87 \\
Before & 2,030 & 0.81 \\
\hline
\end{tabular}}
\vspace{-2mm}
\caption{Causal and temporal relations extracted.}
\vspace{-5mm}
\label{tab:extracted_relations}
\end{table}

\vspace{1mm}
\paragraph{TCAG Visualization}

\begin{figure}
  \centering 
  \includegraphics[scale=0.5]{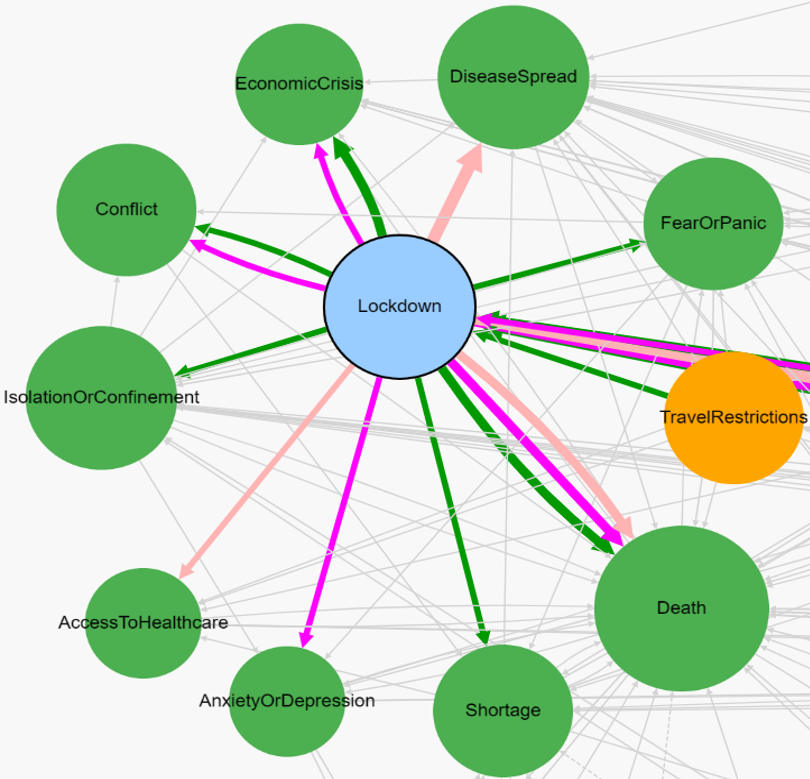}
  \vspace{-7mm}
  \caption{A screenshot of a partial TCAG centered on Lockdown. Green, pink, and purple edges shows Cause, Mitigate and Before relations, respectively. Blue, orange and green nodes show the focused node and nodes with incoming and outgoing edges (with respect to the focused node), respectively.}
    \vspace{-5mm}
  \label{fig:tcag}
\end{figure}

We developed an interactive visualization of the TCAG. Figure~\ref{fig:tcag} shows a small part of the TCAG centered on the event Lockdown. Each node represents an event type in our COVID event taxonomy for which Excavator is able to extract events and track their popularity scores (\S ~\ref{event_timeline}) through time. The three types of relational edges (Causes, Mitigates and Before) are shown in different colors. The size of the nodes and the thickness of the edges indicate the relative frequency of the event types or relations in the log scale, respectively. For example, Figure~\ref{fig:tcag} shows that Death is mentioned more frequently than Lockdown, and the causal relation \{Lockdown, Causes, EconomicCrisis\} appears more frequently than \{Lockdown, Mitigates (``reduces''), AccessToHealthcare\}. To support analysis focusing on a single event, we color the focused event in blue, events that cause or precede the focused event in orange, and events that the focused event causes or precedes in green. 

  \vspace{-2mm}
\paragraph{Event popularity timeseries visualization}

For each node (event) in the TCAG visualization, we show its event popularity timeseries visualization on the side. Figure~\ref{fig:trendall} shows 3 screenshots of the event popularity timeseries (\S ~\ref{event_timeline}) visualization between January and May 2020 for Lockdown, EconomicCrisis and COVID-19 respectively.

\vspace{-1mm}
\section{Recommended Use Cases}

We describe 3 recommended use cases below. More details are in our demonstration video. 
  \vspace{-3mm}
\paragraph{Use case 1: causal and temporal analysis.} We can get a panoramic view of the underlying casual and temporal dynamics between events related to COVID from the overall TCAG. We can start by analyzing the causal or temporal relations centered at an event of interest. For example, Figure~\ref{fig:tcag} shows a diverse range of effects and consequences of Lockdown, such as EconomicCrisis (economic), Shortage (supply-chain), FearOrPanic (mental), etc. Interestingly, the graph also reveals surprises such as \{Lockdown, Causes, Death\}: the UI shows supporting evidence such as ``lockdown exacerbates deaths and chronic health problems associated with poverty, ...''.  Furthermore, the TCAG shows that Lockdown mitigates DiseaseSpread but it also has a negative impact on the Economy, which will inform the decision makers that they will need to understand the economic trade-offs when implementing the Lockdown policy.

We can also analyze longer-distance causal pathways consisting of two or more causal/temporal edges. For example, our demo video shows that COVID-19 causes or precedes (Before) Lockdown, and that Lockdown causes or precedes EconomicCrisis. This helps us understand details about how COVID causes EconomicCrisis. 

\begin{figure*} [t]
  \centering 
  \includegraphics[scale=0.675]{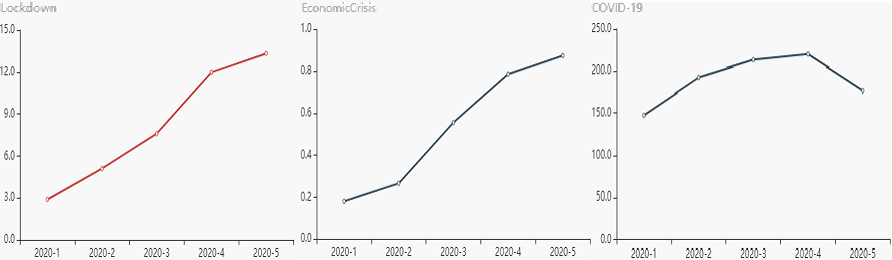}
    \vspace{-6mm}
  \caption{Event popularity timeseries (timeline) between 01-05/2020 for Lockdown, EconomicCrisis and COVID-19. X-axis shows months between January and May 2020. Y-axis shows event popularity scores. }
  \vspace{-4mm}
  \label{fig:trendall}
    \vspace{-2mm}
\end{figure*}

  \vspace{-3mm}
\paragraph{Use case 2: trend and correlation analysis.} We can inspect the event timeline for a node or an edge to perform a trend analysis and a correlation analysis, respectively. Figure~\ref{fig:trendall} shows screenshots of the event popularity timeseries between January and May 2020 for Lockdown, EconomicCrisis and COVID-19. First, the user can click on a single event to perform a trend analysis: the popularity of Lockdown goes up continuously, indicating an upward trend in implementing lockdown policies in more geographic regions. The user can also click on a edge to perform a correlation analysis for a pair of events: when the user clicks on the edge \{Lockdown, Causes, EconomicCrisis\}, the UI shows a strong correlation between the two upward curves. For another edge ``Lockdown mitigates COVID-19'', the UI shows a negative correlation near the end: as Lockdown rises, COVID-19 slightly falls towards the end. 

  \vspace{-3mm}
\paragraph{Use case 3: analyses targeted at geolocations.}

\begin{figure}[h]
  \centering 
    \vspace{-1mm}
  \includegraphics[scale=0.6]{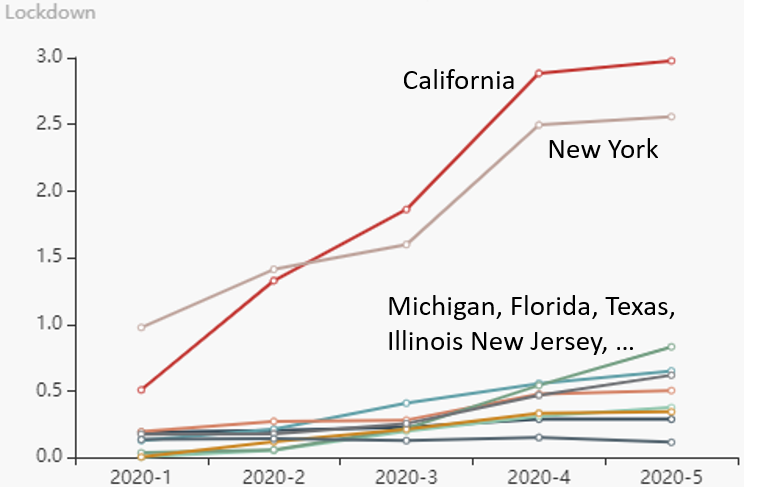}
  \vspace{-7mm}
  \caption{Event popularity timeseries in 01-05/2020 for Lockdown, for top-10 frequently mentioned US states. }
    \vspace{-4mm}
  \label{fig:trendstates}
\end{figure}

The event timeline visualization also allows the user to see the timeline for geolocations such as each U.S. state individually, instead of the aggregate for the entire U.S.. Figure~\ref{fig:trendstates} is a screenshot showing the 10 timelines for Lockdown for the top-10 most frequently mentioned U.S. states. The screenshot shows that the curves for California and New York go much higher than other states. This roughly matches the stricter lockdown policies implemented in the two states during this time period, comparing to other states. Such targeted analysis is made possible because our events have {\it location} and {\it time} arguments. We can also make the TCAG only show events and relations for a specific state, if a user selects a state of interest in the UI.

\section{Related Work}
  \vspace{-2mm}
\paragraph{Extracting events. } Event extraction has been studied using feature-based approaches ~\cite{Huang:2012:MTC:2900929.2900964, ji-grishman:2008:ACLMain}, or neural networks ~\cite{chen-EtAl:2015:ACL-IJCNLP2, nguyen-cho-grishman:2016:N16-1, wadden-etal-2019-entity, liu-etal-2020-event}. GDELT~\cite{leetaru2013gdelt} creates an event database for the conflict and mediation domain. It has very few event types related to COVID-19. To adapt event extraction to new domains, 
Chen et al. ~\shortcite{chan-etal-2019-rapid} developed a user-in-the-loop rapid event customization system. 
% , which we used in Excavator for curating training examples for event extraction. 
Nguyen et al. \shortcite{nguyen-EtAl:2016:RepL4NLP} proposed a neural model for event type extension given seed examples.  Peng et al. \shortcite{PengSoRo16} developed a minimally supervised approach using triggers gathered from ACE annotation guideline.

  \vspace{-2mm}
\paragraph{Extracting causal and temporal relations.} There are a lot of work in 
temporal ~\cite{d2013classifying,chambers-etal-2014-dense,ning-etal-2018-improving, meng-rumshisky-2018-context, han-etal-2019-deep, vashishtha-etal-2020-temporal, wright-bettner-etal-2020-defining} and
causal~\cite{bethard-martin-2008-learning,do2011minimally,riaz-girju-2013-toward, roemmele-gordon-2018-encoder, hashimoto-2019-weakly} relation extraction. Mirza and Tonelli \shortcite{mirza2016catena} and Ning et al. \shortcite{ning-etal-2018-joint} extract both in a single framework. 

  \vspace{-2mm}
\paragraph{Constructing Causal Graphs from Text. } Eidos~\cite{sharp-etal-2019-eidos} uses a rule-based approach to extract causal relations to build a causal analysis graph, that has limited coverage on events related to COVID-19. LearnIt~\cite{min2020learnit} enables rapid customization of causal relation extractors. LearnIt does not focus on causal relations involving COVID-related events. This work also differs from these two in that we extract event arguments and temporal relations, and track event popularity.

  \vspace{-2mm}
\section{Conclusion}
\label{sec:conclusion}
  \vspace{-2mm}
We present the Excavator system, a web-based TCAG visualization, and a video demonstration.

\section*{Acknowledgments}

This research is based upon work supported in part by the 
Office of the Director of National Intelligence (ODNI),
Intelligence Advanced Research Projects Activity (IARPA),
via Contract No.: 2021-20102700002. The views and conclusions 
contained herein are those of the authors and should not be 
interpreted as necessarily representing the official policies, 
either expressed or implied, of ODNI, IARPA, or the U.S.
Government. The U.S. Government is authorized to reproduce 
and distribute reprints for governmental purposes not 
withstanding any copyright annotation therein.

% This document does not contain technology or technical data controlled under either the U.S. International Traffic in Arms Regulations or the U.S. Export Administration Regulations.

\bibliographystyle{acl_natbib}
\bibliography{anthology,acl2021}

%\appendix

\end{document}